\documentclass[11pt,american]{article}

\usepackage[T1]{fontenc}
\usepackage[latin9]{inputenc}
\usepackage[a4paper]{geometry}
\usepackage{amsmath}
\usepackage{amssymb}
\usepackage{graphicx}
\usepackage{setspace}
\usepackage{wasysym}
\usepackage[authoryear]{natbib}
\makeatletter
\newcommand{\lyxaddress}[1]{
  \par {\raggedright #1 
  \vspace{1.4em}
  \noindent\par}
}

\usepackage[T1]{fontenc}

\makeatother

\usepackage{babel}
\begin{document}
\title{Competitive plasticity to reduce the energetic costs of learning}
\author{Mark C.W. van Rossum}
\maketitle

\lyxaddress{School of Psychology and School of Mathematical Sciences, University
of Nottingham, Nottingham NG7 2RD, United Kingdom}
\begin{abstract}
The brain is not only constrained by energy needed to fuel computation,
but it is also constrained by energy needed to form memories. Experiments
have shown that learning simple conditioning tasks already carries
a significant metabolic cost. Yet, learning a task like MNIST to 95\%
accuracy appears to require at least $10^{8}$ synaptic updates. Therefore
the brain has likely evolved to be able to learn using as little energy
as possible. We explored the energy required for learning in feedforward
neural networks. Based on a parsimonious energy model, we propose
two plasticity restricting algorithms that save energy: 1) only modify
synapses with large updates, and 2) restrict plasticity to subsets
of synapses that form a path through the network. Combining these
two methods leads to substantial energy savings while only incurring
a small increase in learning time. In biology networks are often much
larger than the task requires. In particular in that case, large savings
can be achieved. Thus competitively restricting plasticity helps to
save metabolic energy associated to synaptic plasticity. The results
might lead to a better understanding of biological plasticity and
a better match between artificial and biological learning. Moreover,
the algorithms might also benefit hardware because in electronics
memory storage is energetically costly as well.
\end{abstract}

Energy availability is a vital necessity for biological organisms.
The nervous system is a particularly intensive energy consumer. Though
it constitutes approximately 2\% of human body mass, it is responsible
for roughly 20\% of basal metabolism -- continuously consuming some
20W \citep{Attwell:2001cs,Harris2012}. Energy requirements of neural
signaling are now widely seen as an important design constraint of
the brain. Sparse codes and sparse connectivity (pruning) can be used
to lower energy requirements \citep{Levy1996}. Learning rules have
been designed that yield energy efficient networks \citep{Sacramento2015,https://doi.org/10.48550/arxiv.2103.06562}.

However, learning itself is also energetically costly: in classical
conditioning experiments with flies the formation of long-term memory
reduced lifespan by 20\% when the flies were subsequently starved
\citep{Mery2005b}. Moreover, starving fruit flies halt long-term
memory formation; nevertheless forcing memory expression reduced their
lifespan \citep{Placais13}. We have estimated the cost of learning
in fruitflies at 10mJ/bit \citep{maxime_Eestimate}. In mammals there
is physiological evidence that energy availability gates long lasting
forms of plasticity \citep{Potter2010}. Furthermore, in humans there
is behavioral evidence for a correlation between metabolism and memory
\citep{smith2011glucose,Klug2022}.

Given this evidence, we hypothesize that biological neural plasticity
is designed to be energy efficient and that including energy constraints
could lead to computational learning rules more closely resembling
biology, and a better understanding of observed synaptic plasticity
rules. As precise metabolic cost models are not yet available, we
aim at algorithmic principles rather than biophysical mechanisms that
increase energy efficiency. Specifically we examine how energy needed
for plasticity can be reduced while overall learning performance can
be maintained. 

Our study is inspired by experimental observations that despite common
fluctuations in synaptic strength, the number of synapses undergoing
permanent modification appears restricted. First plasticity is spatially
limited, e.g. during motor learning plasticity is restricted to certain
dendritic branches \citep{Cichon2015}. Second, plasticity is temporally
limited. E.g. it is not possible to induce late-phase LTP twice in
rapid succession \citep{Rioult-Pedotti2000,kramar2012synaptic}. 
This stands in stark contrast with traditional backprop which updates
\emph{every} synapse on \emph{every} trial which, as we shall see,
can lead to very inefficient learning.

We use artificial neural networks as model of neural learning. While
neural networks trained with back-propagation are an abstraction of
biological learning, it allows for an effective way to teach networks
complex associations that is currently not matched by more biological
networks or algorithms. Biological implementations of back-propagation
have been suggested \citep[e.g.][]{sacramento2018dendritic}, but
it is likely that these are less energy efficient as learning times
are typically longer. 

In artificial neural network research there is a similar interest
in limiting plasticity, but typically for other reasons. Carefully
selecting which synapses to modify can prevent overwriting previously
stored information, also known as catastrophic forgetting \citep{Sezener2021}.
Randomly switching plasticity on and off can help regularization,
as in drop-out and its variants \citep{salvetti1994introducing}.
Communication with the memory systems is also energetically costly
in computer hardware. Strikingly, storing two 32bit numbers in DRAM
is far more expensive than multiplying them \citep{han2016eie}. Recent
studies have started to explore algorithms that reduce these costs
by limiting the updates to weight matrices \citep{han2016eie,golub2018full}.

\section*{Methods}

\subsubsection*{Data-set}

As data set we use the standard MNIST digit classification task. Similar
results were found on the fashionMNIST data set. The data were offset
so that the mean input was zero. This common pre-processing step is
important. Namely, when trained with backprop, the weight update to
weight $w_{ij}$ between input $x_{j}$ to a unit with error $\delta_{i}$
is
\[
\Delta w_{ij}=-\epsilon W\delta_{i}x_{j}
\]
As the weight update is proportional to the input value $x_{j}$,
there is no plasticity for zero-valued inputs, even when $\delta_{i}\neq0$.
However, this could be a confounding factor as we would like to be
in full control of the number of plasticity events. After zero-meaning
the inputs, only a negligible number of inputs will be exactly zero
and this issue does not arise. Yet, the relative amount of achievable
savings is not substantially affected by this assumption, Fig.~\ref{fig:control}.

\subsubsection*{Network architecture}

We used a network with 784 input units (equal to the number of MNIST
pixels) with bias, a variable number of (non-convolutional) hidden
units in a single hidden unit layer, and 10 output units (one for
each digit). Networks had all-to-all, non-convolutional connectivity.
The hidden layer units used a leaky rectifying linear activation function
(lReLU), so that $g(x)=x$ if $x\geq0$ and $g(x)=\beta x$ with $\beta=0.1$.
A rectified linear activation function $g(x)=\max(0,x)$ would lead
to a substantial fraction of neurons with zero activation in the hidden
layer on a given sample and turn off plasticity of many synapses between
hidden and output layer. As above, this would potentially be confounding,
but it is avoided by using `leaky` units. This did not substantially
change the achievable savings, Fig.~\ref{fig:control}.

\subsubsection*{Training}

The activation of the ten output units is used to train the network.
The target distribution was one-hot encoding of the image labels.
For classification tasks it is common to apply a soft-max non-linearity
$y_{i}=\exp(h_{i})/\sum_{j}\exp h_{j}$, where $h_{i}$ is the net
input to each unit, so that the output activities represent a normalized
probability distribution. One then trains the network by minimizing
the cross entropy loss between the output distribution and the target
distribution. 

However, it is easy to see that when the target is a deterministic
one-hot (0 and 1) distribution, the cross entropy loss only vanishes
in the limit of diverging $h_{i}$, which in turn requires diverging
weights. As cross-entropy loss minimization is energy inefficient,
we used linear output units and minimized the mean squared error between
output and target. This did not substantially change the achievable
savings, Fig.~\ref{fig:control}.

We used the backpropagation learning algorithm to train the network.
Weights were initialized with small Gaussian values ($\sigma=0.01$).
 In preliminary simulations, regularization was seen to have no significant
effect on energy or savings, and was subsequently omitted. Samples
were presented in a random order. Because batching would require storing
synaptic updates across samples, which would be biologically challenging,
the training was not batched and the synaptic weights were updated
after every sample. We did not find substantial energy savings when
using adaptive learning schemes such as ADAM. As such schemes would
furthermore be hard to implement biologically, we instead fixed the
learning rate.

Every 1000 samples training was interrupted and performance on a validation
data set was measured. As a reasonable compromise between performance
and compute requirements, networks were always trained until a validation
set accuracy criterion of 95\% was achieved.

\subsection*{Estimating the energy needed for plasticity}

Our approach requires a model of the metabolic energy needed for learning.
It is currently not known why the metabolic cost of some forms of
plasticity is so high, nor is it known which process is the main consumer.
In flies persistent, protein synthesis dependent memory is much more
costly than protein synthesis independent forms of memory. One could
presume that protein synthesis itself is costly, however it has also
been argued that protein synthesis is relatively cheap \citep{karbowski2019metabolic}.
After all, also in the absence of synaptic plasticity there is substantial
protein turnover. Examples of other costly processes could be transport
of proteins to synapses, actin thread-milling, synaptogenesis. In
addition, there might be energy costs that are not related to individual
synaptic plasticity, such as replay processes (see Discussion). 

We propose a generic model for the metabolic energy $M$ of synaptic
plasticity under the following assumptions: 

1) While it is well known that there are interactions between plasticity
of nearby synapses \citep[e.g.][]{Harvey2007,wiegert2018fate}, we
assume that there is no spatial interaction in regards to synaptic
costs. It might be that potentiation of two synapses is relatively
cheaper than potentiation of a single synapse, e.g. when costly pathways
can be shared, as in synaptic tagging and capture \citep{Frey1997b,Barrett2009,Redondo2011}.
However, instead it is also possible that there is competition for
resources \citep{Fonseca2004,sajikumar2014competition,Jeong2021},
and potentiation of two nearby synapses could be extra costly \citep{Triesch2018}.
The interactions when one synapse undergoes potentiation while another
one undergoes depression, might even be more complicated as resources
might be reused. In summary, both cooperative as well as competitive
interactions likely exist but as there is currently no information
about their energetics, we need to ignore them. The This implies that
the energy is the sum of the individual costs, $M=\sum_{i}M^{(i)}$,
where $i$ sums over all weights in the network, and $M^{(i)}$ is
the energy required to modify synapse $i$. 

2) Similar arguments can be made about the temporal interactions between
plasticity. Again, there is extensive literature on temporal interactions
in plasticity induction and expression \citep{petersen98,abraham02,kramar2012synaptic}.
Energetically, it might be cheaper to potentiate a synapse twice in
rapid succession, but it could equally be more costly. We assume that
there is no temporal interaction in regards to synaptic costs. This
implies that the total metabolic cost is summed over all timesteps
$M=\sum_{t}M(t)$.

\begin{figure}

\includegraphics[width=10cm]{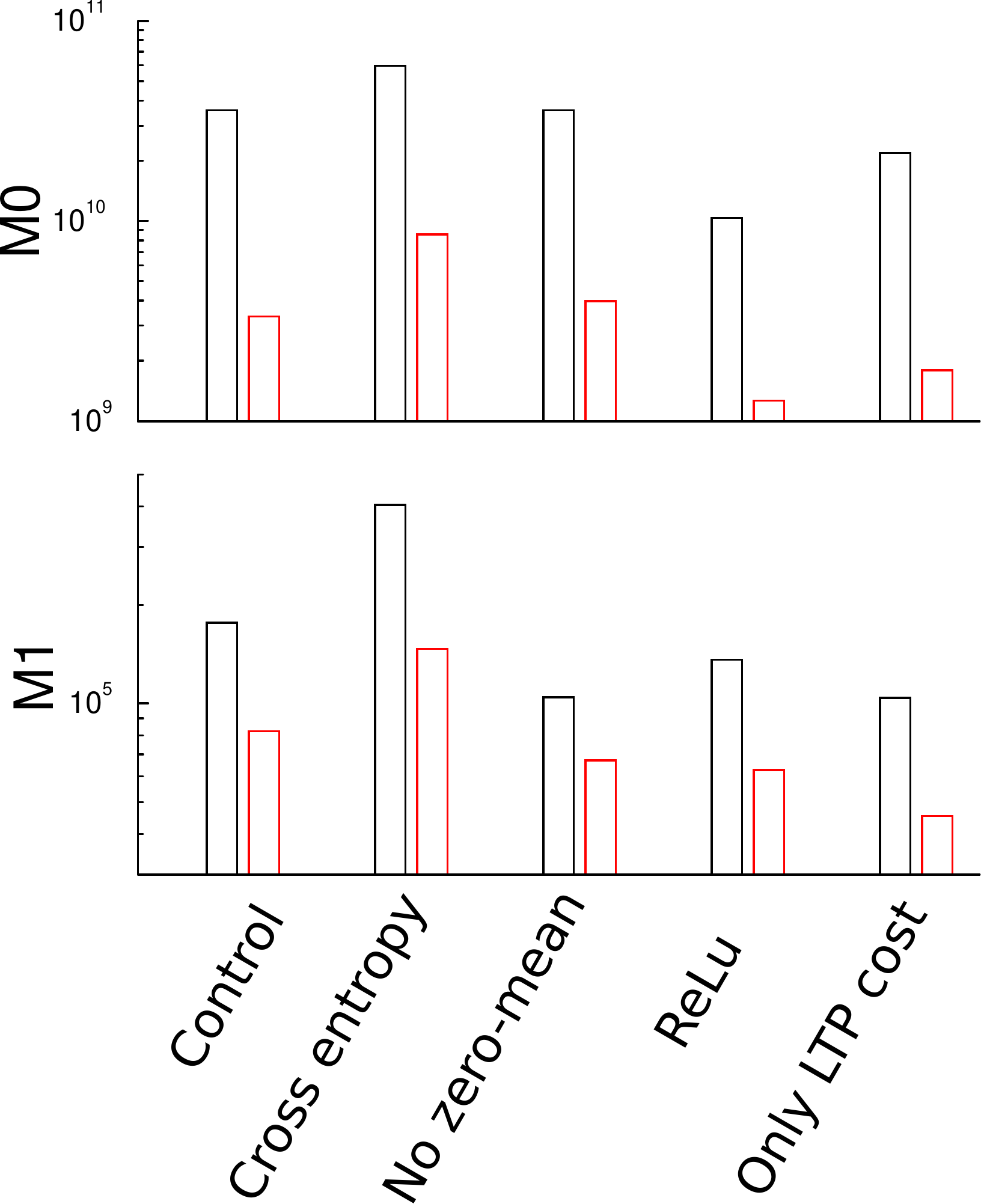}\caption{\label{fig:control}Influence of various assumptions on energy. Each
pair of bars shows the energy required for a network with 2500 hidden
units with unconstrained plasticity (black), and when using the optimal
fraction of plastic synapses with a fixed mask (red). From left to
right, \textbf{Control:} as in main text for comparison; \textbf{Cross
entropy: }training on cross entropy loss function; \textbf{No zero-mean:
}without zero-meaning the data; \textbf{ReLU: }using linear rectifying
(ReLU) units in the hidden layer; \textbf{Only LTP cost:} only positive
weight changes cost energy, negative changes are free. While the energy
levels change, the amount of savings achievable remains similar.}
\end{figure}

3) Next, we assume that synaptic potentiation and depression both
incur the same cost. As we are typically interested in savings in
energy between algorithms, this assumption turns out to be minor.
Consider a case where, say, synaptic potentiation costs energy but
depression does not. As synapses undergo a similar number of potentiation
and depression events during training such a variant would halve the
cost, however, it does not substantially change the achievable savings,
Fig.~\ref{fig:control}.

Under the assumption of spatial and temporal independence we still
require an expression how much modification of a single synapse costs.
We propose that this scales as $|\delta w_{i}|^{\alpha}$, where $\alpha$
is a parameter. The total energy sums across all neurons and all training
time steps is
\[
M_{\alpha}=\sum_{i,t}|\delta w_{i}(t)|^{\alpha}
\]
The parameter $\alpha$ expresses the proportionality in the weight
change. For $\alpha=1$ the energy ($M_{1}$) is linear in the amount
of weight change and represents the accumulated weight changes. This
is for instance relevant for the energy consumed by protein synthesis
where larger changes would require more proteins. Larger values of
$\alpha$ would lead to a situation where updating a synapse twice
would be cheaper than updating once with double the amount. This is
not impossible, but seems unlikely.

In the limit $\alpha\rightarrow0$, $M_{\alpha}$ counts the total
number of synaptic updates, irrespective of their size. As an example,
this would be the case when synaptic tag setting would be costly \citep{Barrett2009}.
It is also a reasonable cost function for digital computer architectures,
where the cost of writing a memory is independent of its value. While
intermediate values of $\alpha$ as well as combinations of terms
are possible, we concentrate on $M_{0}$ and $M_{1}$.

The energy measures are summed across both layers of the network.
The number of synapses of both layers is proportional to the number
of hidden units. But as dictated by the task, the input-to-hidden
layer has 784 synapse per hidden unit, while the hidden-to-output
layer has only 10 synapses per hidden unit. In unmodified networks
the plasticity cost of the input-to-hidden layer therefore dominates
the total energy.

\section*{Results}

\subsection*{Energy of learning}

We first explored the energy requirements of training standard large
networks. We compared the training of a network with 100, 1000, and
10000 hidden units, Fig.\ref{fig:standardnet}a. In addition to the
number of iteration needed to reach criterion performance (top panel),
we measured the two energy variants described in the Methods. The
total number of synaptic updates, called $M_{0}$, is directly proportional
to the learning time. If the learning rate is too low, learning takes
too long; too high and learning fails to converge. An intermediate
learning rate minimizes the learning time and hence $M_{0}$ energy,
Fig.\ref{fig:standardnet}a; middle panel. The optimal learning rate
is in good approximation independent of network size.

The cumulative weight change, called $M_{1}$ energy, measures the
total amount of absolute weight changes of all synapses across training,
Fig.\ref{fig:standardnet}a; bottom panel. This energy measure is
smallest in the limit of small learning rates, where it becomes independent
of the learning rate. With a small learning rate the path in weight
space is more cautious without overshooting associated to larger learning
rates. While the energy measures $M_{0}$ and $M_{1}$ thus have a
different optimal learning rate, below we use a learning rate of 0.01
as a compromise, allowing direct comparison.

\subsection*{Cost of learning in large networks}

\begin{figure}
\includegraphics[width=16cm]{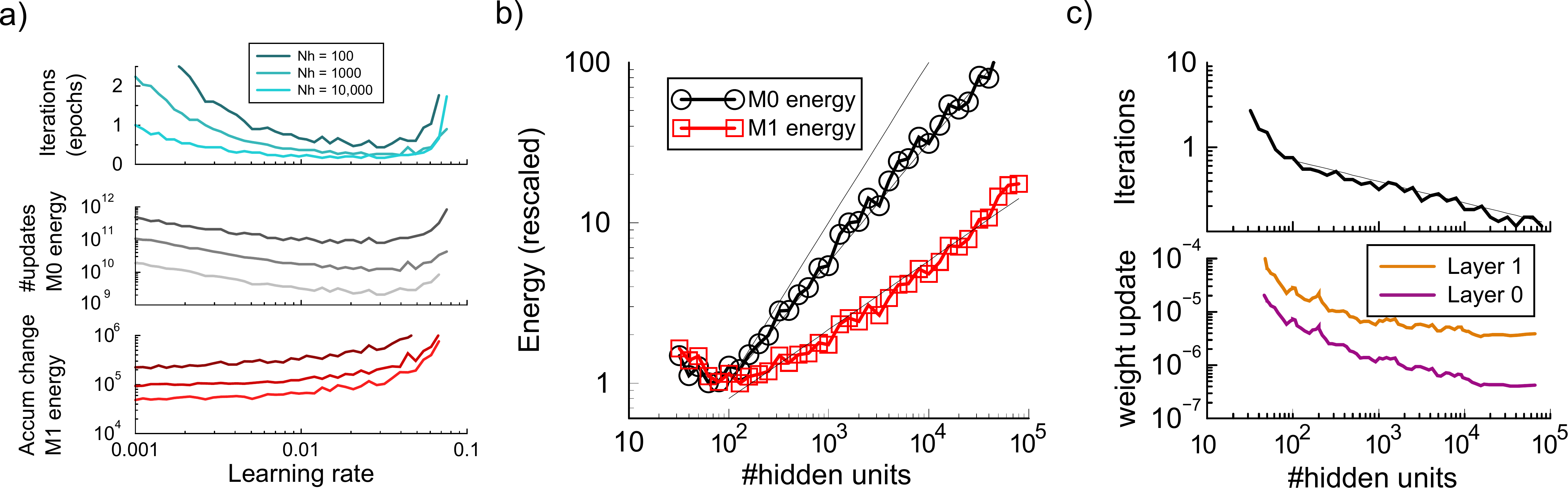}

\caption{\textbf{Energy requirements to train large standard backpropagation
networks. }a). Energy required to train a network with 100, 1000,
or 10000 hidden units as a function of the learning rate. Top panel:
number of iterations to reach 95\% accuracy. Middle panel: The total
number of updates ($M_{0}$ energy) is minimal when the number of
iterations is minimal. Bottom panel: The accumulated changes ($M_{1}$
energy) is lowest in the limit of small learning rates.\protect \\
b). Energy measures versus network size. Since the network has one
hidden layer, network size is expressed in the number of hidden units.
Black curve is the number of updates ($M_{0})$, red is the sum of
all absolute changes ($M_{1}$); both grow in large networks. The
y-axis was scaled so that the minimum energy was one. The three thin
lines represent slope one and powerlaw fits (see text).\protect \\
c). Learning speed and update size vs network size. Top: Large networks
train somewhat faster to the 95\% accuracy criterion than small networks.
Bottom: the mean absolute weight update is smaller in large networks.
\label{fig:standardnet}}
\end{figure}

The MNIST classification networks typically have layers consisting
of some hundred units. However, as in biology the number of neurons
is enormous (see Discussion), we examine the energy consumed as a
function of the network size, Fig.\ref{fig:standardnet}b. First consider
the total number of updates, $M_{0}$. Because our setup rules out
synaptic updates that are accidentally exactly zero (see Methods),
the number of non-zero updates is the total number of synapses in
the network multiplied by the number of iterations $T$. Larger networks
learn a bit quicker than smaller ones, $T\propto N_{h}^{-0.249}$,
Fig.\ref{fig:standardnet}c top, where $N_{h}$ denotes the number
of units in the hidden layer. As a result, energy scales sub-linearly
with network size ($M_{0}\propto N_{h}T\propto N_{h}^{0.75}$). 

The accumulated synaptic change energy $M_{1}$ also increases for
larger networks albeit less steeply, Fig.\ref{fig:standardnet}b.
Despite a fixed learning rate, the size of individual weight updates
is smaller in larger networks, Fig.\ref{fig:standardnet}c, bottom
\citep[c.f.][]{Lee2019}. In large networks individual synapses 1)
take fewer and smaller steps, and 2) the final individual weights
have smaller norm as the larger parameter space of large nets allows
solutions closer to the initial weights. We find an approximate square
root relation between energy and size, $M_{1}\propto N_{h}^{0.43}$,
Fig.\ref{fig:standardnet}b.

\subsection*{Randomly restricting plasticity to save energy}

The above result shows that while large, over-dimensioned networks
learn a bit faster, training them uses far more energy -- restricting
plasticity might save energy. To examine this we first only allow
plasticity in a random subset of synapses using a random binary mask.
Mask elements were drawn from a Bernoulli distribution. The mask was
fixed throughout training. When a different random mask was used at
every trial \citep[as in][]{salvetti1994introducing}, learning simply
slowed down and masking did not save energy; also using a different
mask for each output class did not save energy.

\begin{figure}
\includegraphics[width=17cm]{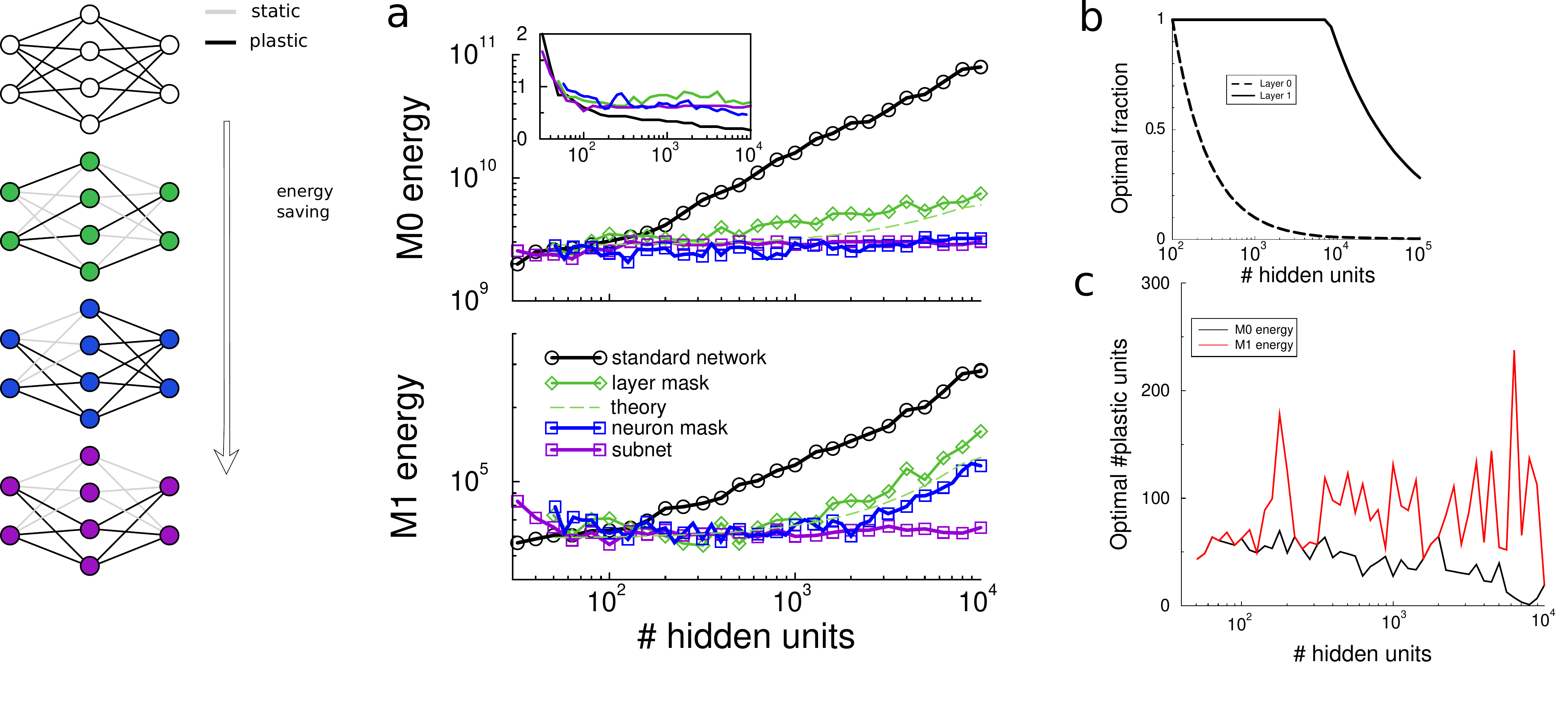}

\caption{Restricting plasticity using random masks saves energy. \protect \\
a) M0 energy (total number of updates) and M1 energy (accumulated
changes) for a standard network against network size (black). A random
synaptic mask saves energy but cannot prevent an eventual increase
in large networks (green; theory in dashed green). A bit more energy
is saved when only a random subset of neurons in the hidden layer
is plastic (blue). Most energy is saved when input and output plasticity
is coordinated so that neurons with plastic inputs have plastic outputs,
energy becomes independent of network size (purple).\protect \\
a-right) Diagram of the various masking settings.\protect \\
b) Theoretically optimal number of plastic synapses (used for green
curve in panel a).\protect \\
c) The optimal number of plastic neurons found when using a neuron
mask (corresponding to blue curve in a).\label{fig:Restricting}}
\end{figure}

Using a mask, both energy measures strongly reduced, Fig.~\ref{fig:Restricting}a,
green curves. The optimal mask density was estimated from a counting
argument. (We also numerically optimized the fraction of plastic synapses
in the input layer; this gave similar results). It is based on the
above observations that to achieve the criterion performance one needs
at least about $100$ hidden units in case that all connections are
plastic, Fig.~\ref{fig:standardnet}b. In other words, there need
to be about $\mu n_{in}n_{out}$ plastic paths between any input and
any output, where $n_{in}=28^{2}$ is the number of inputs and $n_{out}=10$
the number of output units, and $\mu=100$. We now assume that this
is the critical property of the network. According to the hypergeometric
distribution, random masks in the input and output layer reduce the
mean number of plastic paths to $\mu=f_{0}f_{1}N_{h}$, where $f_{i}$
denotes the fraction of plastic connections in layer $i$.

At the optimal probability, we find that the weight change is approximately
the same across network size and layer. Hence both the $M_{0}$ and
$M_{1}$ energy are proportional to $m(f_{0},f_{1})=n_{in}N_{h}f_{0}+n_{out}N_{h}f_{1}$,
We minimize $m$, subject to $\mu=f_{0}f_{1}N_{h}$ and $0\leq f_{i}\leq1$.
For a given $\mu\geq N_{h}$, the energy is minimal when $f_{0}^{*}=\frac{\mu}{N_{h}}\max(1,\sqrt{\frac{N_{h}n_{out}}{\mu n_{in}}})$
and $f_{1}^{*}=\min(1,\sqrt{\frac{\mu n_{in}}{N_{h}n_{out}}})$, see
Fig.\ref{fig:Restricting}b. Because the number of inputs by far exceeds
the number of outputs, it is optimal to keep all hidden-to-output
connections plastic ($f_{1}=1$). Only when $N_{h}\gtrsim\mu n_{in}^{2}/n_{out}\approx8000$,
it becomes wasteful to keep all outgoing connections plastic and it
is best to reduce the fraction of plastic output synapses as well.
The energy scales as 
\begin{equation}
m(f_{0}^{*},f_{1}^{*})=\mu n_{in}\max\left(1,\sqrt{\frac{N_{h}n_{out}}{\mu n_{in}}}\right)+N_{h}n_{out}\min\left(1,\sqrt{\frac{\mu n_{in}}{N_{h}n_{out}}}\right)\label{eq:mscale}
\end{equation}

The energy according to this estimate is plotted as the dashed curve
in Fig.~\ref{fig:Restricting}, with the proportionality constant
extracted from the value of the energy at $N_{h}=100$. The energy
increases in large networks as not every plastic connection in the
input-to-hidden layer is met with a plastic connection in the hidden-to-output
layer. For large networks the energy scales as $m\propto\sqrt{N_{h}}$. 

A similar approach is to mask on a per neuron basis and only allow
plasticity in a fixed subset of neurons in the hidden layer. In practice
we multiply the vector of back-propagated errors element-wise with
a fixed binary mask. The mask density was optimized for each network
size and for both energy measures separately. This algorithm saved
a bit more energy but in large networks the lack of coordination between
plasticity input and output layers, again wastes energy Fig.\ref{fig:Restricting}a,
blue curves. For the $M_{1}$ energy the optimal number of plastic
units in the hidden layer is around 50...100, irrespective of network
size, Fig. \ref{fig:Restricting}c. In contrast the $M_{0}$ energy
was minimal at a lower number of plastic neurons, preferring fewer,
but larger synaptic updates.

There is a trade-off between energy saving and learning speed. In
unrestricted networks, training is faster is large networks, but unsurprisingly,
with all masking variants, training no longer speeds up, Fig.\ref{fig:Restricting}a
inset.

It is interesting to note that for networks with many hidden units
($N_{h}\apprge5000$), it is possible to fix all synapses in the hidden
layer, and restrict plasticity to the hidden-to-output synapses, so
called extreme learning machines \citep{schmidt1992feed,Huang2006}.
However, the plasticity in the hidden-to-output synapses still requires
energy and we found that when trained with SGD such a setup did not
save more energy than the methods presented here. Moreover, it required
a precise tuning of the variance of the input-to-hidden weights.

\subsection*{Coordination of plasticity: subnets}

The above algorithm does save a lot of energy compared to standard
backprop but very large networks still require more energy. Inspired
by the masking algorithm, we coordinate plasticity between input and
output of the neuron: if and only if neuron's incoming connections
are plastic, then so are its outgoing ones. For large networks this
is saves energy because such coordination ensures that energy becomes
independent of network size, Fig.\ref{fig:Restricting}a, purple diagram+curves.
This has a straightforward explanation: the coordination of incoming
and outgoing plasticity results in a plastic sub-network embedded
in a larger fixed network. The energy needed for plasticity is the
same as in a network with the size of the subnet, hence the presence
of the static neurons do not hinder or facilitate the learning. The
optimal size of the plastic subnet weakly depends on the energy measure
used (about 60 for $M_{0}$ and 100 for $M_{1}$). As can be inferred
from Fig.\ref{fig:Restricting}a, using, say, 1000 neurons for the
subnet does not dramatically increase energy. 

\subsection*{Restricting plasticity to synapses with large updates}

In an effort to further decrease energy need, we next modified only
synapses with the largest magnitude updates \citep{golub2018full}.
Upon presentation of a training sample, the proposed weight updates
in the input-to-hidden layer were calculated via standard back-propagation.
However, only synapses with the largest magnitude updates (positive
or negative) were competitively selected and modified. Synapses in
the hidden-to-output layer were always updated. The set of selected
synapses was recalculated for every iteration. 

\begin{figure}
\begin{centering}
\includegraphics[width=0.9\linewidth]{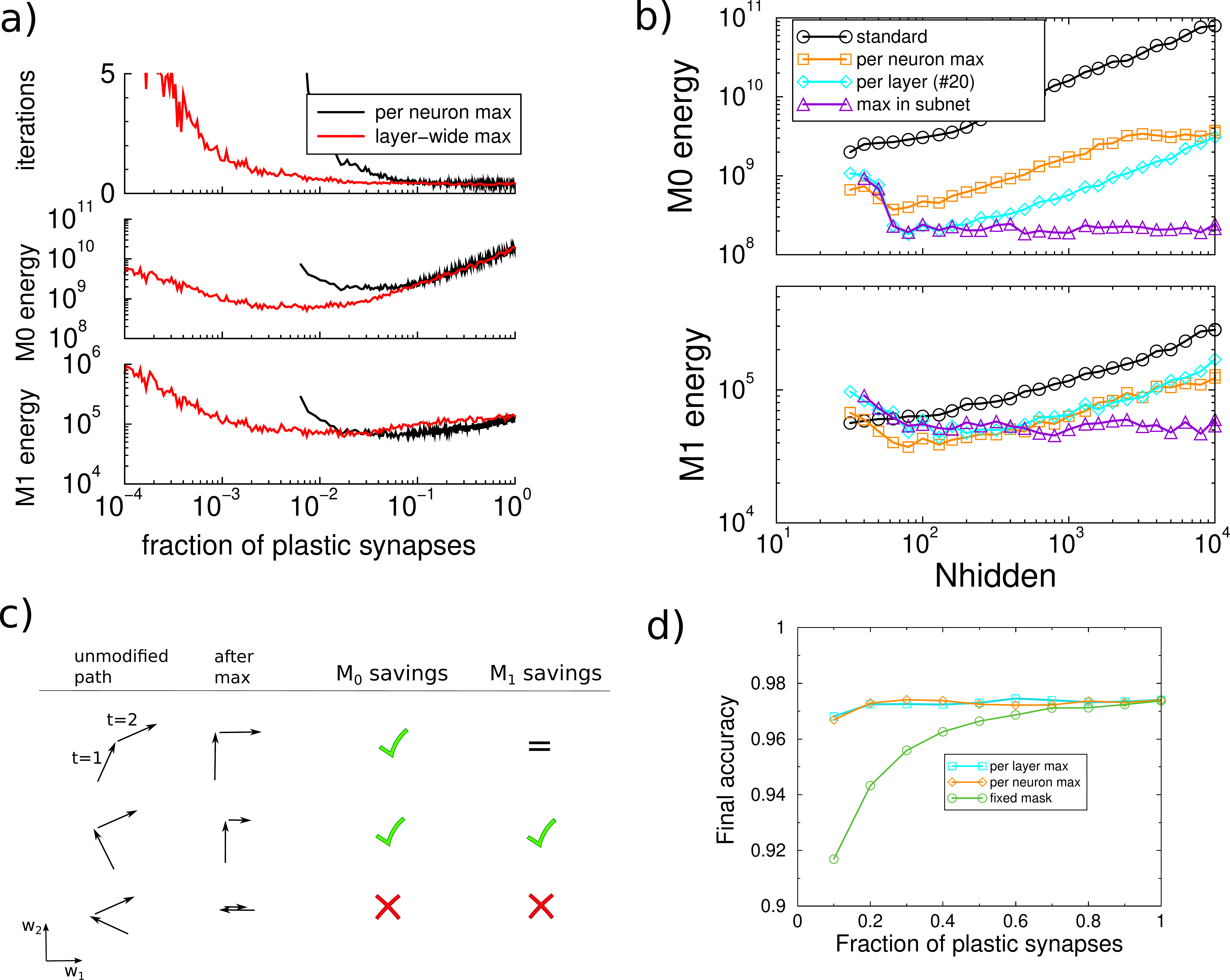}
\par\end{centering}
\caption{Energy required for network training when only synapses with large
updates are modified. a) Iterations and energy in a network with 1000
hidden units as a function of the fraction of plastic synapses. The
largest updates were selected across the layer (red) or for each hidden
layer unit (black).\protect \\
b) Energy requirements as a function of the network size. Allowing
only the largest synaptic updates per layer saves substantial amounts
of energy (cyan curve), as did only allowing the largest updates per
neuron (orange curve), compared to a standard network where all synapses
are updated (black curve). The fraction of updated synapses was optimized
for each network size. Further savings can be achieved by restricting
the selection to subnets (purple), leading to a size independent energy
need. \protect \\
c) Diagram of two subsequent weight updates in three hypothetical
case (top to bottom). After the maximum selection, the updates are
along the cardinal axes. In a particular when the weight trajectory
does not return on itself (top), the selection saves $M_{0}$ energy
but $M_{1}$ energy is identical. \protect \\
d) Test accuracy of a small network (100 hidden units) trained to
saturation. A fixed mask limits performance, but the maximum selection
does not. \label{fig:max}}
\end{figure}

The competition was done in two ways: 1) on the level of each neuron
so that only fraction of synapses per neuron was plastic, 2) across
the whole input-to-hidden layer. It is also possible to select only
neurons with the largest activities, or neurons with the largest backpropagation
error to have plastic synapses. This also saves energy, but not as
much as selecting on weight update per layer. The savings are illustrated
in Fig.~\ref{fig:max}a for a network with 1000 hidden units. Restricting
the synaptic updates to those with the largest gradients saves $M_{0}$
energy in particular, Fig.~\ref{fig:max}a.

We wondered if this competitive algorithm in effect works the same
as the fixed mask. In other words, does it always select the same
synapses to be updated? To examine this we calculated the probability
that a given synapses is updated throughout training and extract the
inverse Simpson index $q=1/[N\sum_{i}p_{i}^{2}]$ \citep{Hill1973},
where $p_{i}$ is the extracted update probability for synapse $i$.
When always the same $k$ synapses would be updated, $q=k/N$. When
the updates would be distributed over all synapses, $q=1$. We find
that $q\approx0.6$ both at the start and end of training, which is
much larger than $k/N\approx0.01$. In contrast to the fixed masks
used above, plasticity keeps switching between synapses under this
algorithm.

Fig. \ref{fig:max}b examines the savings across network size, showing
large $M_{0}$ but little $M_{1}$ savings. The reason for this can
be seen in the diagram Fig.~\ref{fig:max}c. The $M_{1}$ energy,
i.e. L1 path length, is independent of the weight trajectory when
it does not backtrack (top row). Only when the weight paths backtrack
can $M_{1}$ be saved (middle row), but even this is not guaranteed
(bottom row). 

Next, we further sought to decrease energy need by selecting only
the synapses with the largest updates in the subnet. This is the same
algorithm as above but now restricted to the subnet. This further
decreased the energy needs and again made them independent of network
size, Fig.\ref{fig:max} (purple curve).

In summary, the most energy efficient plasticity occurs when 1) on
a given sample only the largest updates are implemented, and 2) these
synapses form a plastic pathways in the network. The competitive selecting
via the maximum has another advantage over a fixed mask. While the
fixed mask limits the capacity of the network, the maximum selection
does not. To show this we used a small network (100 hidden units)
and trained for 30 epochs, at which time accuracy had saturated. When
using a fixed mask, the restricted network's performance drops with
mask size, Fig.\ref{fig:max}d. However using either competitive selection
mechanism, the maximum accuracy dropped much less.

\subsubsection*{}

\subsubsection*{Combining mask with synaptic caching}

\begin{figure}
\begin{centering}
\includegraphics[width=0.9\linewidth]{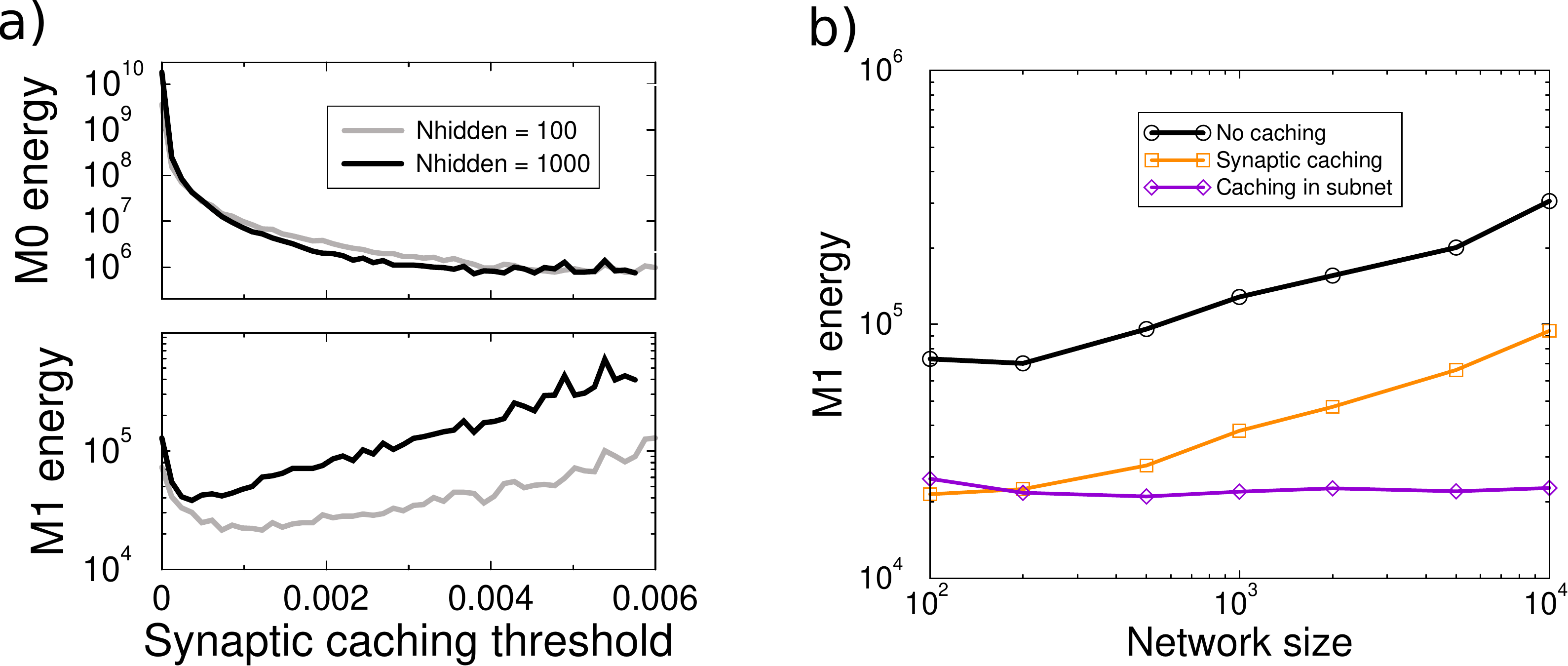}
\par\end{centering}
\caption{\textbf{Synaptic caching. }a) Energy savings achieved by synaptic
caching for networks with 100 and 1000 hidden units. The $M_{0}$
energy that only counts consolidation events, is minimal when the
threshold is large, and is independent of network size. The $M_{1}$
energy trades off cost transient plasticity with consolidation costs
and learning time.\protect \\
b) Synaptic caching saves $M_{1}$ energy. The $M_{1}$ energy is
made independent of size by restricting the plasticity to subnets.
\label{fig:Synaptic-caching} }

\end{figure}
It might appear that we have exhausted all ways to save plasticity
energy, but additional savings can be obtained by reducing the inefficiency
across trials. During learning the synaptic weights follow a random
walk, often partly undoing the update from the previous trial. This
is inefficient.  

An additional, different way to save energy consumed by plasticity
exploits that not all forms of plasticity are equally costly. Transient
changes in synaptic strength (early phase LTP in mammals, ARM in fruit-flies)
are metabolically cheap, while in contrast permanent changes (late-phase
LTP in mammals, LTM in fruit-flies) are costly \citep{Mery2005b,Placais13,Potter2010}.
 It is possible to save metabolic cost by storing updates initially
in transient forms of plasticity and only occasionally consolidate
the accumulated changes using persistent plasticity. We have termed
this saving algorithm \emph{synaptic caching} \citep{li2020energy}.
It is somewhat similar to batching algorithms in machine learning. 

By distributing plasticity over transient and persistent forms, synaptic
caching can save energy. The amount of savings depends on how rapid
transient plasticity decays relatively to the sample presentations,
and its possible metabolic costs. The largest saving can be achieved
when transient plasticity comes at no cost and decays so slowly that
all consolidation can be postponed to a single consolidation step
at the end of learning.

To explore whether the algorithms introduced in this study are compatible
with synaptic caching we implemented both transient and persistent
forms of plasticity in the network coded in $w^{\textrm{trans}}$
and $w^{\textrm{pers}}$. The total connection strength between neurons
was $w^{\textrm{trans}}+w^{\textrm{pers}}$. Updates of the weights
were first stored in the transient component. Only when the absolute
value of the transient component reached a threshold value, the transient
weight was added to the persistent weight and the transient component
was reset to zero. The decay rate of the transient weights was $10^{-3}$
per sample presentation; the cost of transient plasticity was set
to $M_{1}^{\textrm{trans}}=c\sum_{i}|w_{i}^{\textrm{trans}}|$, with
constant $c=0.01$ (see \citealp{li2020energy} for motivation and
further parameter exploration). 

Fig.~\ref{fig:Synaptic-caching} shows the energy measures as a function
of the consolidation threshold. Synaptic caching saves a substantial
amount of energy. As the transient plasticity does not contribute
to the $M_{0}$ energy measure, it now just counts the number of synaptic
consolidation events. It is lowest at a high consolidation threshold,
at even higher thresholds ($\apprge0.06$) the learning no longer
converges. By thus limiting and postponing consolidation the $M_{0}$
energy with synaptic caching is virtually independent of network size. 

However, the $M_{1}$ energy still increases with network size, with
a similar dependence on network size as the standard network, Fig.~\ref{fig:Synaptic-caching}b.
It will still lead to high costs in large networks. Coordinating both
transient and persistent plasticity, that is restricting plasticity
to subnets, again eliminates this increase. We also tried a variant
in which transient plasticity was distributed over the whole network
and only consolidation was coordinated, however this did not save
energy.

In sum, combining synaptic caching with the above savings strategy
saves the most energy. For the $M_{0}$ energy synaptic caching even
completely obviates the need for additional saving strategies in large
networks. We emphasize however that synaptic caching does rely on
additional synaptic complexity, namely two synaptic plasticity components
and a consolidation mechanism. Moreover, if the updates of the transient
component would also incur $M_{0}$ energy, additional saving strategies
might be possible.

\section*{Discussion}

\begin{figure}
\begin{centering}
\includegraphics[width=8cm]{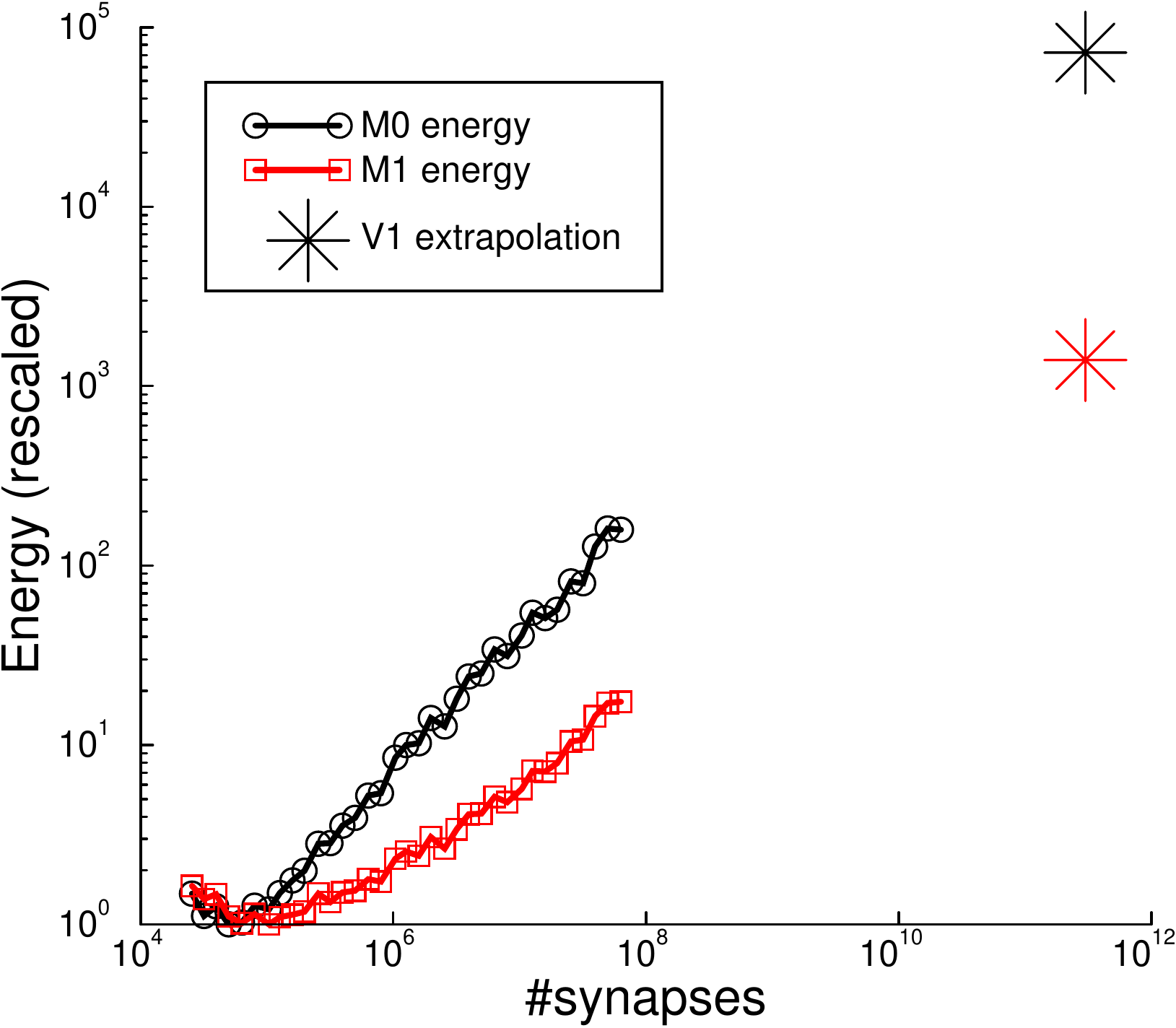}
\par\end{centering}
\caption{Extrapolation of results of Fig.\ref{fig:standardnet}b to the number
of synapses in macaque V1. Naive, unrestricted backprop learning would
use $10^{5}\times$ more synaptic updates and $10^{3}\times$ more
$M_{1}$ energy than minimally required. \label{fig:extrapol}}
\end{figure}

Experiments have shown that already simple associative learning is
metabolically costly. For neural networks that learn more complicated
tasks, energy costs can become very high, in particular when networks
are large. For instance, macaque V1 has some 150 million neurons and
some 300$\times10^{9}$ synapses \citep{Vanni2020}. Extrapolating
Fig.\ref{fig:standardnet}, if plasticity were distributed over all
these synapses, the number of synaptic updates would be some $10^{5}$
times larger than required. The $M_{1}$ energy, which increases less
steeply with network size, would still be some 700 times larger, Fig.\ref{fig:extrapol}.
Thus large networks are powerful, but without restrictions the metabolic
cost of plasticity could be very large. 

We have introduced two approaches to reduce costs. First, restrict
plasticity to a subset of synapses. This is most efficient when the
plasticity on input and output side of a neuron are coordinated, so
that when inputs of a neuron are modified then so are its outputs.
Such effects have indeed been observed, although the precise rules
appear complex \citep{Fitzsimonds1997,Tao2000,Zhang2021}.

Second, express plasticity only in synapses with large updates. This
is also consistent with neuroscience data, where there is typically
a threshold for plasticity induction. Our study would suggests the
presence of a competitive algorithm in which only a certain fraction
of synapses is modified. Biophysically, the competition on the neuron
level could naturally follow from resource constraints.

These strategies can be combined with our earlier work on synaptic
caching. Finally it also possible to skip over uninformative samples
during learning, leading to even further saving \citep{Pache2023}. 

Future studies could include how the saving algorithms should be automatically
adapted dependent on task difficulty or network architecture. Another
interesting avenue is to find the most energy efficient synaptic modification
during transfer learning \citep[cf.][]{Oswald2021}. 

Given the lack of biological data and the uncertain nature of the
main energy consumer, our proposed energy model is currently coarse.
In particular the independence assumption is unlikely to be fully
correct; yet even the sign of the interactions (cooperative vs competitive)
is unknown. We assumed that the energy is extensive in the number
of synapses. It might be that there is a large component independent
of synapse number, for example energy needed for replay processes.
In that case, the question is at what number of synapses the energy
becomes approximately extensive. As more experimental data becomes
available, the energy model can be refined and the efficiency of the
proposed algorithms can be re-examined. 

Finally, we did not consider the energy required for neural signaling,
such as synaptic transmission and spike generation. Ultimately, learning
rules should aim to also reduce those costs.

\subsection*{Acknowledgments}

It is a pleasure to thank Mikhail Belkin, Simon Laughlin, Thomas Oertner,
Aaron Pache, Joao Sacramento, Long Tran-Thanh, and Silviu Ungureanu
for discussion. This research was supported by a grant from NVIDIA
and utilized an NVIDIA RTX A6000. 

\bibliographystyle{plainnat_initialsonly_nomonth}
\bibliography{neuro_jab}

\subsection*{\protect\pagebreak Appendix}

\end{document}